%% file: main.tex
\documentclass{article}

     \PassOptionsToPackage{numbers, compress}{natbib}

\usepackage[preprint]{neurips_2023}

\usepackage[utf8]{inputenc} 
\usepackage[T1]{fontenc}    
\usepackage{hyperref}       
\usepackage{url}            
\usepackage{tabularx}
\usepackage{booktabs}       
\usepackage{amsfonts}       
\usepackage{nicefrac}       
\usepackage{microtype}      
\usepackage{xcolor}         

\usepackage{times}
\usepackage{epsfig}
\usepackage{graphicx}
\usepackage{amsmath}
\usepackage{amssymb}
\usepackage{booktabs}
\usepackage{multirow}
\usepackage{mmstyle}
\usepackage{enumitem}
\usepackage{subcaption}
\usepackage{bm}

\usepackage{wrapfig}

\title{MATLABER: Material-Aware Text-to-3D via \\ LAtent BRDF auto-EncodeR}

\def \shlab{$^\dagger$}

\def \ntu{$^\S$}

\author{%
  Xudong Xu\shlab \qquad Zhaoyang Lyu\shlab \qquad Xingang Pan\ntu \qquad Bo Dai\shlab \\
  \shlab Shanghai AI Laboratory \ \ntu S - Lab, Nanyang Technological University  \\ 
  {\tt\small \shlab \{xuxudong, lvzhaoyang, daibo\}@pjlab.org.cn \ \ntu xingang.pan@ntu.edu.sg}
}

\begin{document}

\maketitle

\input{sections/abs.tex}
\input{sections/intro.tex}

\input{sections/related_work.tex}
\input{sections/method.tex}
\input{sections/experiments.tex}
\input{sections/conclusion.tex}

{\small
	\bibliographystyle{ieeetr}
	\bibliography{main}
}


\end{document}

%% file: sections/abs.tex

\begin{abstract}

Based on powerful text-to-image diffusion models, text-to-3D generation has made significant progress in generating compelling geometry and appearance. However, existing methods still struggle to recover high-fidelity object materials, either only considering Lambertian reflectance, or failing to disentangle BRDF materials from the environment lights. In this work, we propose Material-Aware Text-to-3D via LAtent BRDF auto-EncodeR (\textbf{MATLABER}) that leverages a novel latent BRDF auto-encoder for material generation. We train this auto-encoder with large-scale real-world BRDF collections and ensure the smoothness of its latent space, which implicitly acts as a natural distribution of materials. 
During appearance modeling in text-to-3D generation, the latent BRDF embeddings, rather than BRDF parameters, are predicted via a material network.
Through exhaustive experiments, our approach demonstrates the superiority over existing ones in generating realistic and coherent object materials.
Moreover, high-quality materials naturally enable multiple downstream tasks such as relighting and material editing.
Code and model will be publicly available at  \url{https://sheldontsui.github.io/projects/Matlaber}.

\end{abstract}

%% file: sections/intro.tex

\section{Introduction}
\label{sec:intro}

3D asset creation is imperative for various industrial applications such as gaming, film, and AR/VR.
Traditional 3D asset creation pipeline involves multiple labor-intensive and time-consuming stages \cite{labschutz2011content},
all of which rely on specialized knowledge and professional aesthetic training. 
Thanks to the recent development of generative models,
recent text-to-3D pipelines that automatically generate 3D assets from purely textual descriptions have received growing attention,
due to their rapid advances in generation quality and efficiency,
as well as their potential of significantly reducing the time and skill requirement of traditional 3D asset creation.

\begin{figure*}[t]
  \centering
  \includegraphics[width=0.98\linewidth]{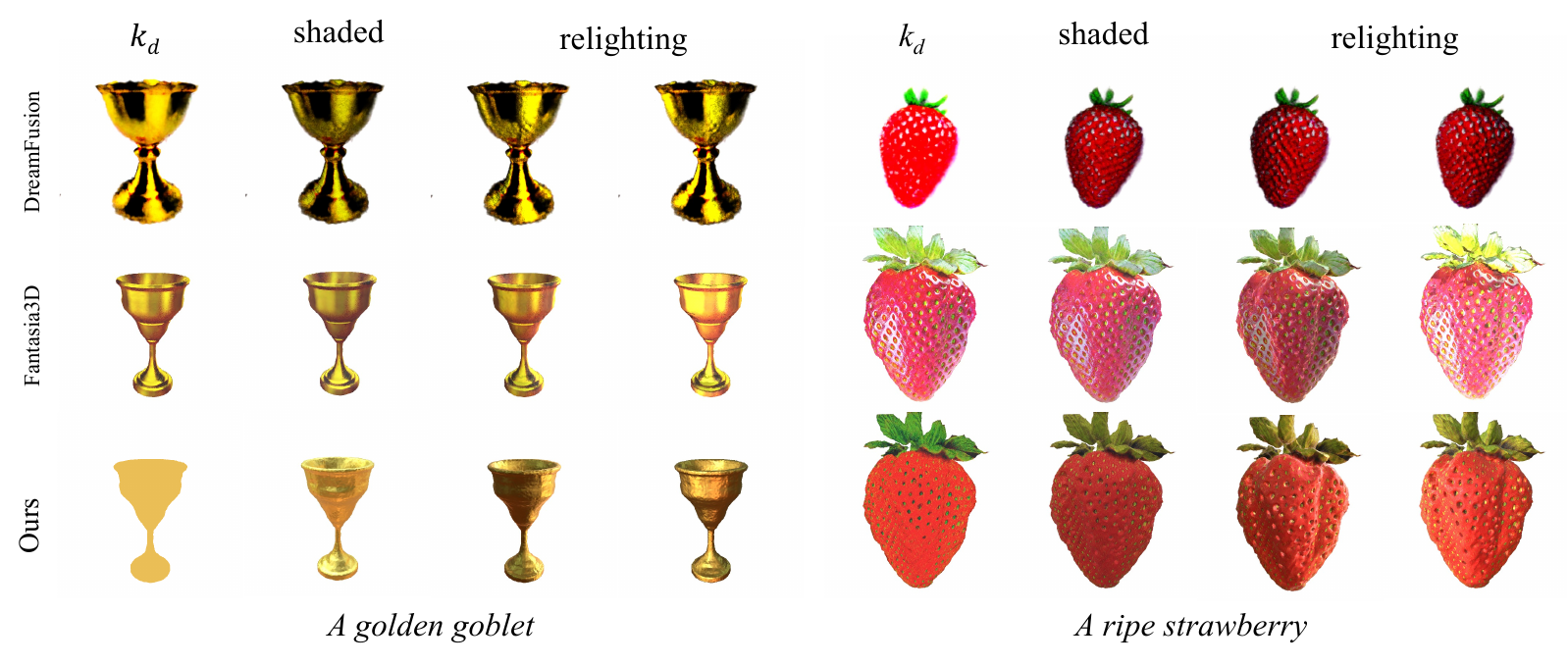}
   \caption{Text-to-3D generation aims to synthesize high-quality 3D assets aligning with given text descriptions. Despite the impressive appearance, representative methods like DreamFusion~\cite{poole2022dreamfusion} and Fantasia3D~\cite{chen2023fantasia3d} still fail to recover high-fidelity object materials. Specifically, DreamFusion only considers diffuse materials while Fantasia3D always predicts BRDF materials entangled with environment lights. Based on a latent BRDF auto-encoder, our approach is capable of generating natural materials for 3D assets, enabling realistic renderings under different illuminations.}
   \label{fig:teaser}
\end{figure*}

Gradually optimizing the target 3D asset represented as  NeRF \cite{mildenhall2021nerf} or \textsc{DMTet}~\cite{shen2021dmtet} through the SDS loss \cite{poole2022dreamfusion},
compelling geometry and appearance can be obtained by these text-to-3D pipelines \cite{jain2022dreamfiled,mohammad2022clip,poole2022dreamfusion,chen2023fantasia3d,lin2022magic3d,wang2023prolificdreamer}.
However, as shown in Figure\ref{fig:teaser}, they still struggle to recover high-fidelity object materials, which significantly limits their real-world applications such as relighting.
Although attempts have been made to model Lambertian reflectance \cite{poole2022dreamfusion,lin2022magic3d}
and bidirectional reflectance distribution function (BRDF) \cite{chen2023fantasia3d},
in their designs, the neural network responsible for predicting materials has no sufficient motivation and clues to unveil an appropriate material that obeys the natural distribution, especially under fixed light conditions,
where their predicted material is often entangled with environment lights. 

In this work, we resort to existing rich material data to learn a novel text-to-3D pipeline that effectively disentangles material from environment lights.
In fact, despite the inaccessibility of paired datasets of material and text descriptions, there exist large-scale BRDF material datasets such as MERL BRDF~\cite{Matusik:2003}, Adobe Substance3D materials~\cite{adobesubstance3d}, and the real-world BRDF collections TwoShotBRDF~\cite{boss2020two}.
Therefore, we propose Material-Aware Text-to-3D via LAtent BRDF auto-EncodeR (\textbf{MATLABER}) that leverages a novel latent BRDF auto-encoder to synthesize natural and realistic materials that accurately align with given text prompts.
The latent BRDF auto-encoder is trained to 
embed real-world BRDF priors of TwoShotBRDF in its smooth latent space, 
so that MATLABER can predict BRDF latent codes instead of BRDF values to focus more on choosing the most suitable material and worry less about the validity of predicted BRDF.

Thanks to the smooth latent space of the BRDF auto-encoder, our approach ensures the realism and coherence of object materials, achieving the ideal disentanglement of geometry and appearance.
As shown in Figure~\ref{fig:teaser}, our approach can create 3D assets with high-fidelity material, outperforming previous state-of-the-art text-to-3D pipelines. 
More importantly,
the effective estimation of object materials naturally allows various operations such as relighting, material editing, and scene manipulation,
which can hardly be achieved before our work.
It's noteworthy that these downstream tasks are extremely crucial for a series of real-world applications, paving the way for a more convenient paradigm of 3D content creation.
Furthermore, by exploiting multi-modal datasets like ObjectFolder, our model has the potential to infer acoustic and tactile information from the obtained materials, which constitute the trinity of material for virtual objects.

%% file: sections/related_work.tex

\section{Related Work}
\label{sec:relwork}

\subsection{Text-to-Image Generation}
In recent years, we have witnessed significant progress in text-to-image generation empowered by diffusion models.
By training on large-scale text-image paired datasets,
diffusion models can implicitly link semantic concepts and corresponding text prompts, and thus is capable of generating various and complex images of objects and scenes~\cite{nichol2021glide,saharia2022imagen,ramesh2022dalle2,rombach2022stablediffusion}.
While GLIDE~\cite{nichol2021glide} obtains text embeddings via a pretrained CLIP~\cite{radford2021learning} model, Imagen~\cite{saharia2022imagen} and eDiff-I~\cite{balaji2022ediffi} adopt larger language models such as T5~\cite{raffel2020exploring} to achieve more diverse image synthesis and a deeper level of language understanding.
To enable text-to-image training on limited computational resources,
Stable Diffusion~\cite{rombach2022stablediffusion} leverages the latent diffusion model (LDM) and trains its diffusion model on the latent space instead of the pixel space, demonstrating highly competitive performance in terms of quality and flexibility.
However, all of these works are constrained to the 2D domain and ignore the huge demand in the 3D field.

\subsection{Text-to-3D Generation with 2D Supervision}
With the great success of text-to-image synthesis, text-to-3D generation has gained significant attention in recent years as a promising way to generate high-quality 3D content from simple textual input.
Different from text-to-image generation, no massive text-3D paired data is available, making it infeasible to train a text-to-3D model as in training a text-to-image model.
Alternatively, DreamField~\cite{jain2022dreamfiled} and CLIP-mesh~\cite{mohammad2022clip} explore the 2D supervision by using a pre-trained CLIP model to optimize the underlying 3D representations, such as NeRFs and meshes, to achieve high text-image alignment scores for multi-view renderings.
Later, DreamFusion~\cite{poole2022dreamfusion} utilizes, for the first time, a powerful 2D text-to-image diffusion model~\cite{saharia2022imagen} as a prior and introduces an effective Score Distillation Sampling (SDS) loss to guide the optimization.
Based on the promising SDS loss,
Magic3D~\cite{lin2022magic3d} optimizes 3D objects in two consecutive stages and further improves the rendered resolution of generated 3D objects from $64$ to $512$, showing impressive performance in text-to-3D generation.
In Fantasia3D~\cite{chen2023fantasia3d}, Chen~\etal proposes to represent 3D objects with flexible \textsc{DMTet} representation~\cite{shen2021dmtet} and model the appearance via the BRDF modeling, which can successfully generate compelling geometry and photorealistic object appearance.
Recently, ProlificDreamer~\cite{wang2023prolificdreamer} proposes a more advanced guiding loss, Variational Score Distillation (VSD), for more diverse and high-fidelity object generation.
Unlike these works, we aim to recover the inherent material information in text-to-3D generation, empowering more downstream applications such as relighting and material editing.

\subsection{Material Estimation}
In the community of computer vision and graphics, researchers aim to estimate surface materials for decades.
The Bidirectional Reflection Distribution function (BRDF) is the most widely used material model, which characterizes how a surface reflects lighting from an incident direction toward an outgoing direction~\cite{nicodemus1965directional,li2018materials}.
One line of research works~\cite{lensch2001image,lensch2003image,gao2019deepbrdf,bi2020neural} targets recovering the reflectance from known 3D geometry and some methods~\cite{aittala2013practical,aittala2015two,thanikachalam2017handheld,hui2017reflectance,kang2018efficient,deschaintre2018single} only focus on BRDF acquisition of 2D planar geometry.
Moreover, the simultaneous acquisition of 3D geometry and materials has also gained a surge of interest and several solid papers~\cite{nam2018practical,barron2014shape,boss2021neural,zhang2021nerfactor,munkberg2022extracting} have already achieved compelling results on in-the-wild scenarios.
Specifically, Neural-PIL~\cite{boss2021neural} and NeRFactor~\cite{zhang2021nerfactor} both leverage BRDF datasets for material priors.
Unlike these methods, we aim to create high-quality surface materials aligning with the provided text prompts.
Concurrent to this work, Fantasia3D~\cite{chen2023fantasia3d} also explores the automatic creation of surface materials and object geometries from given language models.
However, their obtained materials are unluckily entangled with lights and thus intractable to put into new environments.

%% file: sections/method.tex

\section{Method}
\label{sec:method}

We present Material-Aware Text-to-3D via LAtent BRDF auto-EncodeR (\textbf{MATLABER}), aiming for photorealistic and relightable text-to-3D object generation.
In the following, preliminaries on score distillation sampling are first presented in Section~\ref{subsec:sds}.
Then in Section~\ref{subsec:prior}, we review appearance modeling in prior works and analyze their intrinsic deficiencies in the relighting scenarios.
Finally, we introduce the latent BRDF auto-encoder in Section~\ref{subsec:brdfae} and discuss how to incorporate it into material-aware text-to-3D generation in Section~\ref{subsec:T23D}.
An overview of our framework is illustrated in Figure~\ref{fig:pipeline}.

\subsection{Score Distillation Sampling (SDS)}
\label{subsec:sds}

DreamFusion~\cite{poole2022dreamfusion} first achieves high-quality text-to-3D generation with a pretrained text-to-image diffusion model.
It represents the scene with a modified Mip-NeRF~\cite{barron2021mip}, which can produce an image $x = g(\theta)$ at the desired camera pose.
Here, $g$ is a differentiable renderer, and $\theta$ is a coordinate-based MLP representing a 3D volume.
The diffusion model $\phi$ comes with a learned denoising function $\epsilon_\phi(x_t;y,t)$ that predicts the sampled noise $\epsilon$ given the noisy image $x_t$, noise level $t$, and text embedding $y$.
Specifically, to update the scene parameters $\theta$, DreamFusion introduces Score Distillation Sampling (SDS), which computes the gradient as: 
\begin{equation}
    \nabla_\theta \mathcal{L}_\text{SDS}(\phi, x) = 
    \mathbb{E}_{t, \epsilon} \! \! \left[ w(t)(\epsilon_\phi(x_t;y,t) - \epsilon)\frac{\partial x}{\partial \theta} \right],
\end{equation}
where $w(t)$ is a weighting function.
Here, we follow Stable-dreamfusion~\cite{stabledreamfusion} or Fantasia3D~\cite{chen2023fantasia3d} and leverage the publicly available latent diffusion model (LDM) like Stable Diffusion~\cite{rombach2022stablediffusion} as our guidance model.
Therefore, the SDS loss now becomes:
\begin{equation}
\label{eq:sds_high}
    \nabla_\theta \mathcal{L}_\text{SDS}(\phi, x) = 
    \mathbb{E}_{t, \epsilon} \!\! \left[ w(t)(\epsilon_\phi(z_t;y,t) \!-\! \epsilon)\frac{\partial z}{\partial x}\frac{\partial x}{\partial \theta} \! \right].
\end{equation}
Intuitively, SDS loss will push all the rendered images to the high probability density regions given by the pretrained diffusion model~\cite{lin2022magic3d}.
Apart from the rendered RGB images, Fantasia3D~\cite{chen2023fantasia3d} regards the surface normal maps $n$ as special images and shows applying SDS loss on normal maps encourages to generate good object geometry with fine details.

\subsection{Appearance Modeling in Text-to-3D Generation}
\label{subsec:prior}

To employ the SDS loss, prior works adopt different schemes to model the appearance of 3D objects. DreamFusion~\cite{poole2022dreamfusion} leverages a reflectance model similar to \cite{bi2020neural,pan2021shading,boss2021nerd,srinivasan2021nerv} and only considers diffuse reflectance~\cite{lambert1760photometria,ramamoorthi2001signal} while rendering multi-view images.
For each point $\vx$ on a specific ray, the RGB albedo $\vrho$ and volumetric density $\tau$ are predicted via a MLP, and the surface normal $\vn$ is obtained with $\vn = - \nabla_{\vx} \tau / \| \nabla_{\vx} \tau \|$.
After accumulating each normal $\vn$ and albedo $\vrho$ in a ray with volume rendering~\cite{max1995optical},
assuming some point light source with 3D coordinate $\boldsymbol{\ell}$ and color $\boldsymbol{\ell}_\rho$, and an ambient light color $\boldsymbol{\ell}_a$,
DreamFusion employs Lambertian shading to produce a color $\vc$ for each pixel as:
\begin{equation}
    \mathbf{c}=\boldsymbol{\rho} \circ\left(\boldsymbol{\ell}_\rho \circ \max (0, \boldsymbol{n} \cdot(\boldsymbol{\ell}-\vx) /\|\boldsymbol{\ell}- \vx \|)+\boldsymbol{\ell}_a\right).
\end{equation}
Although generating various 3D objects with appealing appearances, DreamFusion fails to model specular reflectance, which is an indispensable term in the material that leads to photorealistic renderings under different illuminations.
Magic3D~\cite{lin2022magic3d} uses more advanced \textsc{DMTet}~\cite{shen2021dmtet} as the scene representation and targets at higher-fidelity 3D models but still adopts Lambertian shading when texturing the object meshes extracted from \textsc{DMTet}.
Hence, their synthesized 3D objects all lack complete material information, significantly limiting their applications in real-world scenarios.

To achieve more photorealistic and relightable rendering, Fantasia3D~\cite{chen2023fantasia3d} introduces the spatially varying Bidirectional Reflectance Distribution Function (BRDF) into text-to-3D generation.
They represent the material with three components~\cite{mcauley2012practical}, namely the diffuse term $\vk_d\in \mathbb{R}^3$, the roughness and metallic term $\vk_{rm} \in \mathbb{R}^2$ containing roughness $k_r$ and metalness factor $m$, as well as the normal variation term $\vk_n \in \mathbb{R}^3$.
According to the convention in~\cite{karis2013real}, the specular term $\vk_s$ is computed with $\vk_s = (1-m) \cdot 0.04 + m \cdot \vk_d$.
For a specific surface point $\vx$ with normal $\vn$ and outgoing view direction $\vomega_o$, the final rendering $L(\vx,\vomega_o)$ can be obtained following the rendering equation~\cite{kajiya1986rendering}:
\begin{align}
    \label{eq:rendering}
    L(\vx,\vomega_o) = \int_{\Omega} L_{i}(\vx,\vomega_{i}) f(\vx, \vomega_{i}, \vomega_{o}; \vk_d, \vk_s) (\vomega_{i} \cdot \vn) \mathrm{d}\vomega_{i},
\end{align}
where incident light $L_i$ comes from the direction $\vomega_{i}$ and BRDF $f$ is related to the diffuse term $\vk_d$ and the specular term $\vk_s$.
However, in Fantasia3D~\cite{chen2023fantasia3d}, the metalness factor $m$ is usually set to $0$ and thus the specular term $\vk_s$ is actually ignored during the appearance modeling.
Moreover, they utilize a fixed HDR environment map with uniform brightness distribution as the environment lights all the time.
Despite its appealing appearance under fixed illuminations, Fantasia3D always predicts materials entangled with environmental lights, which leads to unrealistic renderings under novel lighting conditions.

\begin{figure*}[t]
  \centering
  \includegraphics[width=0.98\linewidth]{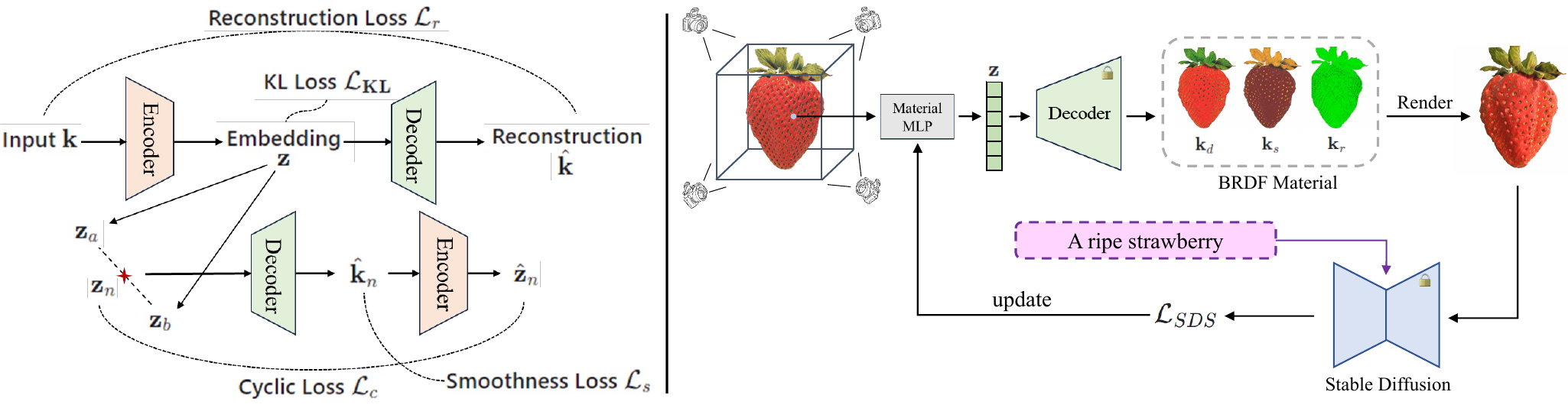}
   \caption{\textbf{Left}: Our latent BRDF auto-encoder is trained on the TwoShotBRDF dataset with four losses, \ie, reconstruction loss, KL divergence loss, smoothness loss, and cyclic loss. Imposing KL divergence and smoothness loss on latent embeddings encourages a smooth latent space~\cite{boss2021neural}. \textbf{Right}: Instead of predicting BRDF materials directly, we leverage a material MLP $\Gamma$ to generate latent BRDF code $\vz$, which is then decoded to 7-dim BRDF parameters via our pretrained decoder. Similar to prior works, the SDS loss can be applied to the rendered images, which empowers the training of our material MLP network. (Note that, roughness $k_r$ is scalar and we visualize it with the green channel in this paper.)}
   \label{fig:pipeline}
   \vspace{-5mm}
\end{figure*}

\subsection{Latent BRDF Auto-Encoder}
\label{subsec:brdfae}

Confronted with the limitations of prior works, we take both diffuse and specular terms into consideration for appearance modeling.
Specifically, for a surface point $\vx$, we aim to estimate the Cook-Torrence~\cite{cook1982reflectance} BRDF parameter $\vk \in \mathbb{R}^7$ including diffuse $\vk_d \in \mathbb{R}^3$, specular $\vk_s \in \mathbb{R}^3$, and roughness $k_r \in \mathbb{R}$.
However, directly optimizing on standard BRDF space lacks necessary constraints and may make the predicted materials fall into invalid BRDF regions.
As mentioned in Section~\ref{sec:intro}, we resort to data-driven BRDF priors that are learned from real-world BRDF data collections.

Inspired by previous literature on reflectance decomposition~\cite{zhang2021nerfactor,boss2021neural},
we leverage TwoShotBRDF~\cite{boss2020two} dataset to train a latent BRDF auto-encoder for such a material prior.
TwoShotBRDF is a real-world BRDF material dataset containing $11,250$ high-quality SVBRDF maps of size $768 \times 768$ collected from various online sources, where each pixel represents an independent BRDF parameter.
Following~\cite{berthelot2018understanding,boss2021neural}, interpolating auto-encoders are adopted here for smoother interpolation in the latent space.
As illustrated in Figure~\ref{fig:pipeline}, our BRDF auto-encoder consists of an MLP encoder $\cE$ and an MLP decoder $\cD$.
Given a BRDF parameter $\vk$, the encoder will generate a 4-dim latent code $\vz = \cE(\vk)$, which is then reconstructed to a 7-dim BRDF $\hat{\vk} = \cD(\vz)$ via the decoder network.
Here, apart from the standard $L_2$ reconstruction loss $\cL_r$, we additionally employ a Kullback–Leibler (KL) divergence loss $\cL_\text{KL} = KL(p(\vz) || \cN(\vzero, \mI))$ to encourage the smoothness of latent space.
Moreover, for an ideal smooth latent space, the linearly interpolated embeddings $\vz_n = \alpha \cdot \vz_a + (1-\alpha) \cdot \vz_b$ between two random latent codes $\vz_a$ and $\vz_b$ could be also decoded to a reasonable BRDF parameter $\hat{\vk}_n = \cD(\vz_n)$.
We, therefore, impose a smoothness loss $\cL_s = \sum_n (\nabla_{\alpha} \cD(\vz_n))^2$ for a batch of uniformly interpolated latent codes $\vz_n$ following~\cite{boss2021neural}.
After adding the $L_2$ cyclic loss $\cL_c$ between interpolated code $\vz_n$ and the re-encoded counterpart $\hat{\vz}_n = \cE(\hat{\vk}_n)$,
the total loss for our latent BRDF auto-encoder training is a combination of four losses:
\begin{align}
    \cL = \cL_r + \lambda_\text{KL} \cL_\text{KL} + \lambda_{s} \cL_{s} + \lambda_{c} \cL_{c},
\end{align}
where $\lambda_\text{KL}$, $\lambda_{s}$ and $\lambda_{c}$ are all balancing coefficients.
With this total loss function, our encoder and decoder networks are trained jointly.


\subsection{Material-Aware Text-to-3D}
\label{subsec:T23D}

In this work, we divide the whole text-to-3D generation into two consecutive stages: geometry generation and appearance generation.
Following Fantasia3D~\cite{chen2023fantasia3d}, we represent the 3D objects with the hybrid scene representation of \textsc{DMTet} owing to its superior performance in geometry modeling and photorealistic surface rendering.
While generating the geometry, we employ the SDS loss on normal maps similar to strategies proposed in~\cite{chen2023fantasia3d}.
For the \textsc{DMTet} predicted by a geometry MLP $\psi$, a differentiable render $g$ can render a normal map $n$ from a randomly sampled camera pose $c$ as follows: $n = g(\psi, c)$.
Then, the SDS loss on normal maps $n$ will help the geometry MLP $\psi$ update until it converges to a satisfactory geometry aligning with the given text prompt.

Once obtaining the geometry of 3D objects, we can leverage our latent BRDF auto-encoder for appearance generation.
As shown in Figure~\ref{fig:pipeline}, for any point $\vx$ on the surface, we first apply the hash-grid positional encoding $\beta(\cdot)$~\cite{muller2022instantngp} and then use a material MLP $\Gamma$ parameterized as $\gamma$ to predict its BRDF latent code $\vz_\vx$, which is then transferred to 7-dim BRDF parameter $\vk_\vx$ via the pre-trained BRDF decoder $\cD$ following:
\begin{align}
    \label{eq:material}
    \vz_\vx = \Gamma(\beta(\vx); \gamma), \quad \vk_\vx = \cD(\vz_\vx).
\end{align}
Equipped with this crucial material information $\vk_\vz = [\vk_d, \vk_s, k_r]$,
the point $\vx$ can be rendered with the aforementioned rendering equation~\ref{eq:rendering} under given incoming light $L_i(\vomega_{i})$ from direction $\vomega_{i}$.
Specifically, for the point $\vx$ with normal $\vn$ and outgoing view direction $\vomega_o$, the final rendering $L(\vx,\vomega_o)$ is the summation of diffuse intensity $L_{d}$ and specular intensity $L_{s}$ as $L(\vx,\vomega_o) = L_{d}(\vx) + L_{s}(\vx,\vomega_o)$, and the two terms can be computed as follows:
\begin{align} \label{eq:split_sum_all}
\begin{split}
     L_{d}(\vx) &= \vk_d (1-m) \int_{\Omega} L_{i}(\vx,\vomega_{i})(\vomega_{i} \cdot \vn) \mathrm{d}\vomega_{i}, \\
     L_{s}(\vx, \vomega_o) &= \int_{\Omega} \frac{DFG}{4(\vomega_o \cdot \vn)(\vomega_{i} \cdot \vn)} L_{i}(\vx,\vomega_{i})(\vomega_{i} \cdot \vn) \mathrm{d}\vomega_{i},
 \end{split}
 \end{align}
where $D$, $G$, and $F$ are functions representing the GGX~\cite{walter2007microfacet} normal distribution (NDF), geometric attenuation, and Fresnel term, respectively.
It's noteworthy that the integration over the hemisphere $\Omega$ can be calculated using the split-sum method following \cite{karis2013real,munkberg2022extracting}.
In particular, the specular intensity $L_s(\vx,\vomega_o)$ is further approximated as: 
\begin{gather}
    F(\vomega_o, \vh, k_r) = F_0 + \big( \max(1-k_r, F_0) - F_0 \big) \big( 1 - (\vomega_o \cdot \vh) \big)^5, \\
    L_s(\vx, \vomega_o) = \big ( F(\vomega_o, \vh, k_r) B_0(\vomega_o \cdot \vn, k_r) + B_1(\vomega_o \cdot \vn, k_r) \big )
    \int_{\Omega} D(\vomega_i, \vomega_o, \vn, k_r) L_i(\vx, \vomega_i) \mathrm{d}\vomega_{i},
\end{gather}
where $\vh = (\vomega_i + \vomega_o) / |\vomega_i + \vomega_o|$ represents the half vector, $F_0$ is the specular albedo $\vk_s$~\footnote{The definition of specular albedo $\vk_s$ is different from that mentioned in Section~\ref{subsec:prior}, which is actually equal to $F(\vomega_o, \vh, k_r)$ computed here. For convenience, we still use the notation $\vk_s$ for specular albedo.}, and $B_0$ and $B_1$ are two pre-computed 2D lookup textures (LUT) indexed by $(\vomega_o \cdot \vn)$ and the roughness $k_r$.

During the appearance modeling, we will leverage multiple environment maps and keep vertically rotating these maps to encourage the predicted BRDF materials to disentangle from environment lights.
Given a sampled camera viewpoint, the rendered image $x$ is actually the aggregation of rendered pixels along the outgoing view direction $\vomega_o$, which can be obtained with Equation~\ref{eq:split_sum_all}.
Accordingly, we can apply the aforementioned SDS loss in Equation~\ref{eq:sds_high} to the rendered image $x$ and the material MLP $\Gamma$ can be optimized with the gradient \wrt its parameter $\gamma$:
\begin{equation}
    \nabla_\gamma \mathcal{L}_\text{SDS}(\phi, x) = 
    \mathbb{E}_{t, \epsilon} \!\! \left[ w(t)(\epsilon_\phi(z_t;y,t) \!-\! \epsilon) \frac{\partial z}{\partial x} \frac{\partial x}{\partial \vk} \frac{\partial \vk}{\partial \gamma} \! \right].
\end{equation}
Apart from the SDS loss, a material smoothness regularizer is extra utilized for enforcing smooth diffuse materials following~\cite{munkberg2022extracting,zhang2021nerfactor}.
For a surface point $\vx$ with diffuse $\vk_d(\vx)$, the regularizer is defined as:
\begin{equation}
    \cL_\text{mat} = \sum_{\vx \in \cS} |\vk_d(\vx) - \vk_d(\vx + \vepsilon)|,
\end{equation}
where $\cS$ denotes the object surface and $\vepsilon$ is a small random 3D perturbation.

%% file: sections/experiments.tex

\section{Experiments}
\label{sec:experiments}

\textbf{Implementation details.}
While training our latent BRDF auto-encoder, we will scatter all the SVBRDF maps to 7-dim BRDF parameters and randomly shuffle them for training. We set the batch size to $256$, and loss balancing coefficients $\lambda_\text{KL}$, $\lambda_{s}$ and $\lambda_{c}$ to 2e-4, $0.05$, and 1e-4, respectively.
Our BRDF auto-encoder is trained for $30$ epochs using the AdamW optimizer with a learning rate of 1e-4 on an NVIDIA A100 GPU.
For the geometry generation, we initialize the \textsc{DMTet} with either a 3D ellipsoid, a 3D cylinder, or a customized 3D model provided by users following~\cite{chen2023fantasia3d}.
During the text-to-3D generation, the 3D object is optimized on 4 NVIDIA A100 GPUs with an AdamW optimizer, where each GPU loads $9$ images rendered from randomly sampled camera poses.
Specifically, our method spends $3,000$ iterations (learning rate $0.001$) on geometry modeling and $2,000$ iterations (learning rate $0.01$) on material generation.
For the weight in score distillation sampling, we always adopt the \emph{strategy I} proposed in Fantasia3D~\cite{chen2023fantasia3d}.

\begin{figure*}[t]
  \centering
  \includegraphics[width=0.98\linewidth]{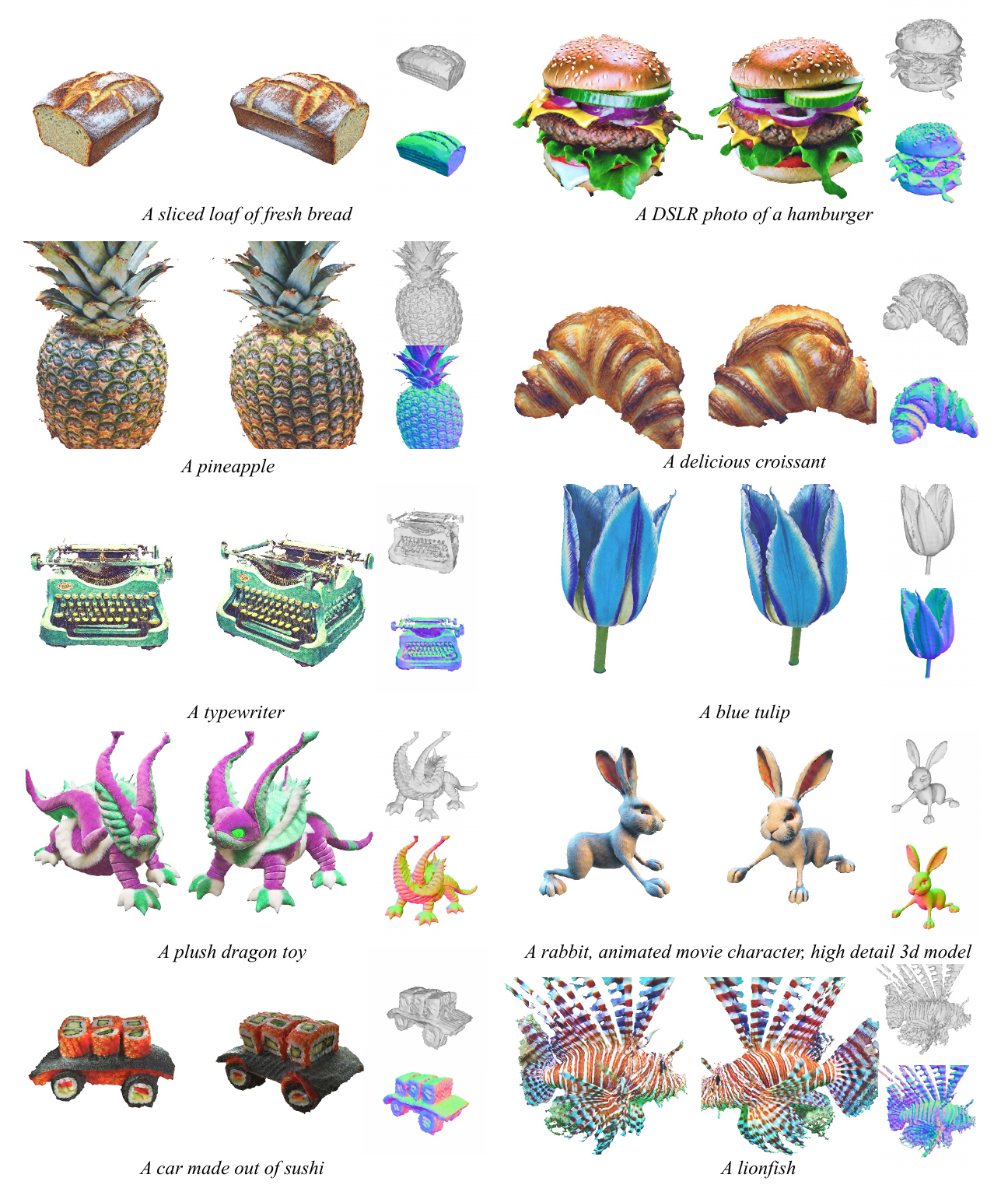}
   \caption{The gallery of our text-to-3D generation results. Shapes, normal maps, and shaded images from two random viewpoints are presented here.}
   \label{fig:quality}
   \vspace{-5mm}
\end{figure*}

\begin{table*}[t]
    \centering
    \caption{Mean opinion scores in range $1 \sim 5$, where 1 means the lowest score and 5 is the highest score.}
		\label{tab:userstudy}
		\begin{tabular}{ccccc}
			\toprule
			  Method & Alignment & Realism & Details & Disentanglement \\
			\midrule
			DreamFusion~\cite{poole2022dreamfusion} & 3.97 (± 0.66) & 3.56 (± 0.43) & 3.23 (± 0.61) & 3.48 (± 0.59) \\
                Magic3D~\cite{lin2022magic3d} & \textbf{4.01} (± 0.59) & 3.84 (± 0.72) & 3.70 (± 0.66) & 3.14 (± 0.89) \\
                Fantasia3D~\cite{chen2023fantasia3d} & 3.76 (± 0.82) & 4.17 (± 0.65) & 4.27 (± 0.75) & 2.93 (± 0.95) \\
                Ours                                 & 3.81 (± 0.75) & \textbf{4.35} (± 0.60) & \textbf{4.31} (± 0.70) & \textbf{3.89} (± 0.65) \\
			\bottomrule
		\end{tabular}
\end{table*}

\begin{figure*}[ht]
  \centering
  \includegraphics[width=0.98\linewidth]{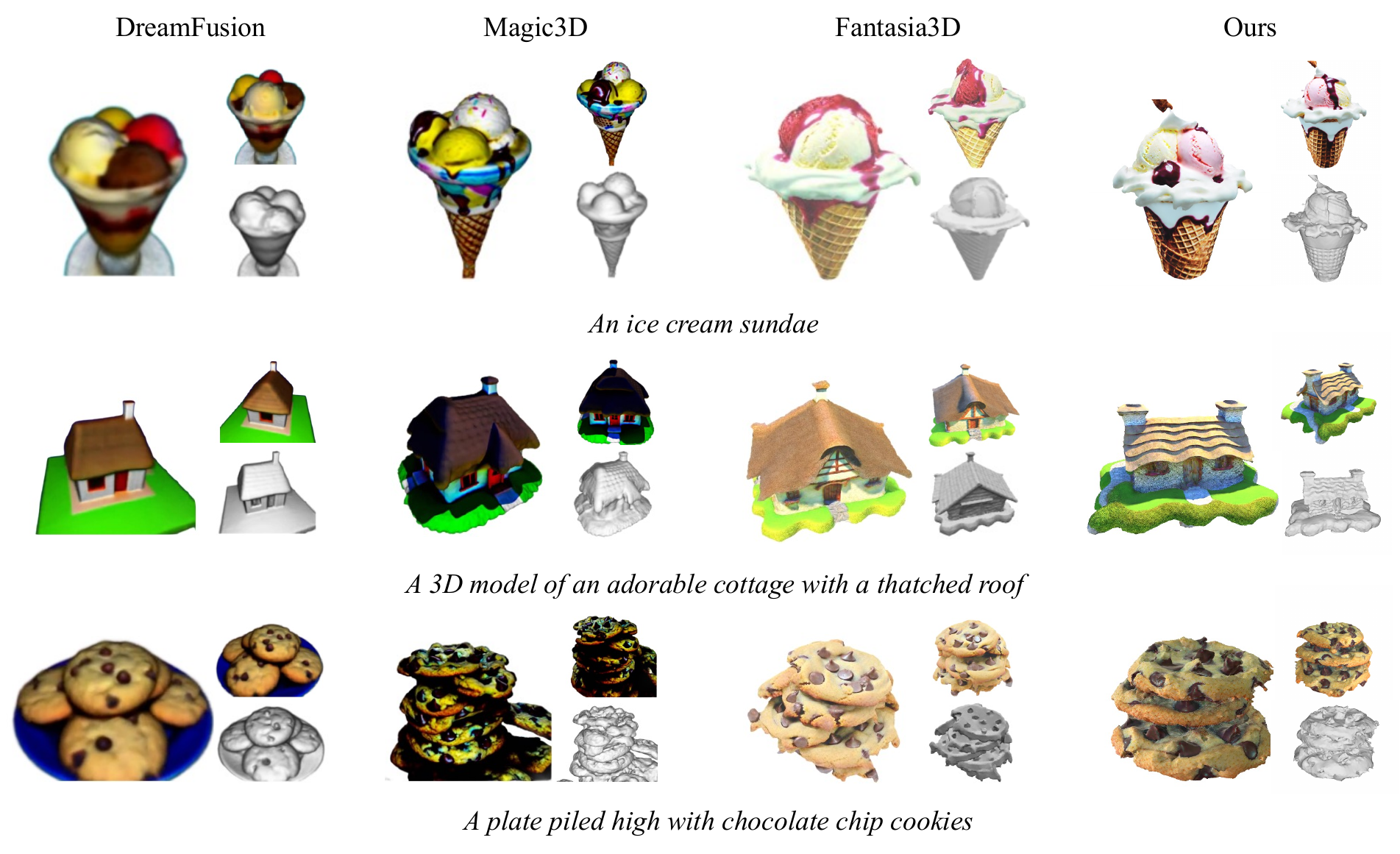}
   \caption{Qualitative comparisons to baselines. Our results have more natural textures and richer details.}
   \label{fig:compare}
   \vspace{-5mm}
\end{figure*}

\begin{figure*}[t]
  \centering
  \includegraphics[width=0.98\linewidth]{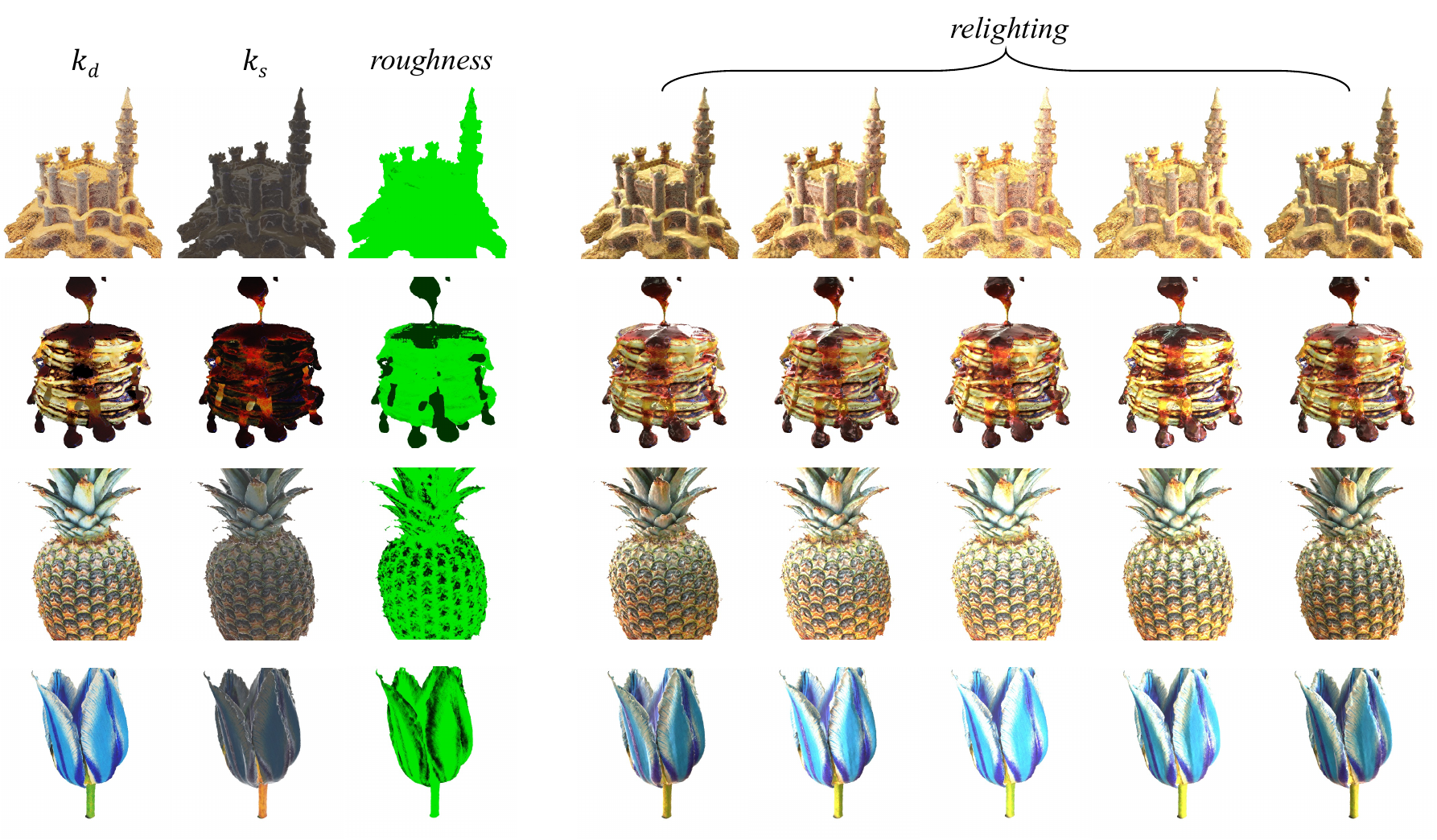}
   \caption{Relighting results. On the left side, we list the generated BRDF materials, including diffuse, specular, and roughness. The relit images under a rotating environment light are presented on the right side.}
   \label{fig:relighting}
   \vspace{-5mm}
\end{figure*}

\begin{figure*}[t]
  \centering
  \includegraphics[width=0.98\linewidth]{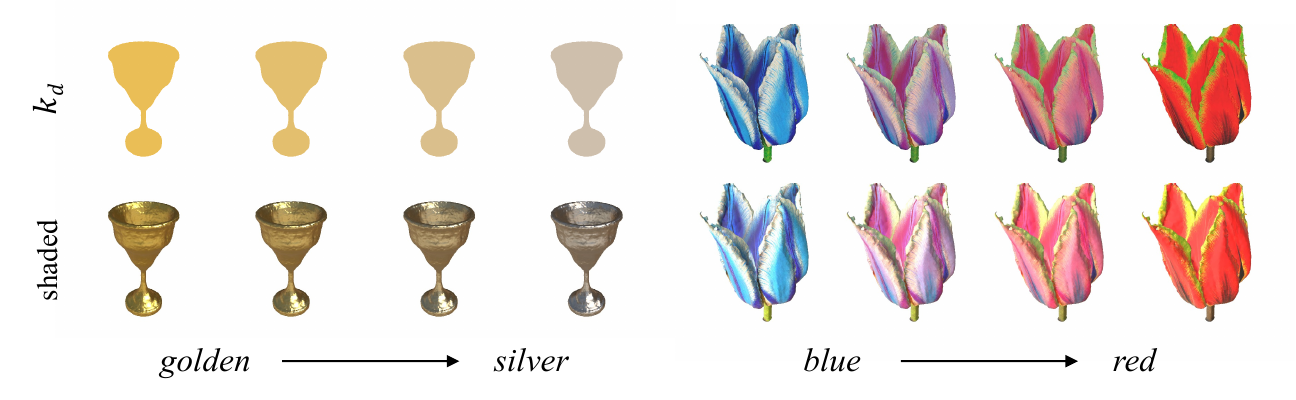}
   \caption{Material interpolation results. Thanks to the smooth latent space of our BRDF auto-encoder, we can conduct a linear interpolation on the BRDF embeddings. As can be observed here, a golden goblet will turn into a silver goblet while the color of the tulip changes from blue to red gradually.}
   \label{fig:interp}
   \vspace{-5mm}
\end{figure*}

\begin{figure*}[h!]
  \centering
  \includegraphics[width=0.98\linewidth]{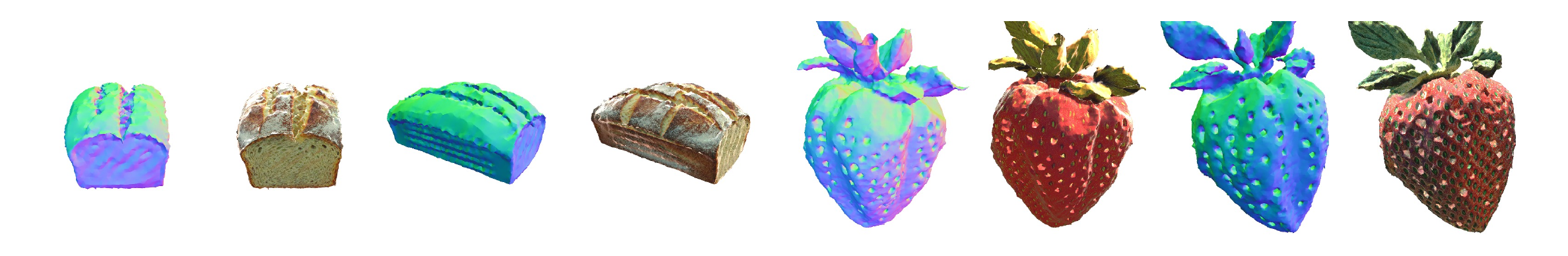}
   \caption{Failure cases. Owing to imperfect geometry, our generated 3D objects will present clear artifacts under some novel illuminations.}
   \label{fig:limitation}
   \vspace{-5mm}
\end{figure*}

\textbf{Qualitative results.}
Given text descriptions, our method MATLABER is capable of generating photorealistic 3D assets and simultaneously obtaining high-fidelity BRDF materials.
In this section, we compare our approach with three representative methods, namely DreamFusion~\cite{poole2022dreamfusion}, Magic3D~\cite{lin2022magic3d}, and Fantasia3D~\cite{chen2023fantasia3d}.
As shown in Figure~\ref{fig:compare}, we present the comparative results given the same text descriptions.
For Fantasia3D, 3D assets are synthesized with their official code, while results of DreamFusion and Magic3D are borrowed from their project pages owing to the inaccessibility of their model weights.
Thanks to the latent BRDF auto-encoder, our method generates competitive geometry as Fantasia3D and demonstrates a better appearance with more natural textures and richer details.
Moreover, the gallery of more 3D assets generated with MATLABER is provided in Figure~\ref{fig:quality}.

\textbf{User study.}
To evaluate the quality of generated 3D objects from human's perspective,
we invited $80$ volunteers to conduct a user study.
For each participant,
he/she will view $10$ randomly selected 3D objects generated by our approach and three baseline methods.
They are asked to evaluate these 3D assets in four different dimensions: `alignment', `realism', `details', and `disentanglement'.
Alignment means how closely the generated 3D objects match the given text prompts, and realism reflects the fidelity of synthesized mesh textures.
Besides, users also need to judge the richness of details on these 3D objects.
Finally, we extra present the corresponding diffuse materials and ask them whether the diffuse is disentangled from the environment lights.

Totally, we collect $800$ responses, and the comparative results are summarized in Table~\ref{tab:userstudy}.
Our method MATLABER achieves the best performance on three evaluation protocols, \ie, realism, details, and disentanglement.
Especially for disentanglement, ours outperforms three baseline methods by a large margin, showing the effectiveness of our proposed latent BRDF auto-encoder model.
In terms of alignment, DreamFusion~\cite{poole2022dreamfusion} and Magic3D~\cite{lin2022magic3d} both capitalize on large-scale language models such as T5~\cite{raffel2020exploring} and therefore demonstrate better alignment with the text descriptions.
Unfortunately, these models are all not released for public usage.

\textbf{Relighting and material editing.}
Since our model is capable of generating BRDF materials, we can implement relighting on these 3D assets.
To validate the fidelity of our generated materials, we manually rotate an HDR environment map and thus obtain a series of HDR maps with different environment lights.
As shown in Figure~\ref{fig:relighting}, the relit objects under different illuminations are all natural and photorealistic.
Moreover, thanks to the smooth latent space of our BRDF auto-encoder, we can also conduct material interpolation if given two different material descriptions.
Figure~\ref{fig:interp} demonstrates a smooth material morphing on the goblet and tulip.
For example, given the text prompts of ``A golden goblet'' and ``A silver goblet'', we can obtain their corresponding material latent embeddings after the optimization, and the final results can be obtained via a naive interpolation on the latent embeddings.

%% file: sections/conclusion.tex

\section{Conclusion}
\label{sec:conclusion}

In this work, we propose MATLABER, a novel latent BRDF auto-encoder for material-aware text-to-3D generation.
This auto-encoder is trained with large-scale real-world BRDF collections, providing implicit material prior for the appearance modeling in text-to-3D generation.
Thanks to such a material prior, our approach can generate high-quality and coherent object materials in text-to-3D synthesis, achieving the ideal disentanglement of geometry and appearance.
Moreover, the generated BRDF materials also support various operations such as relighting, material editing, and scene manipulations.
Meanwhile, our latent BRDF auto-encoder potentially can be used in other tasks to predict materials instead of RGB values to enable the ability of relighting and material editing.

\textbf{Limitations.}
Our generated 3D objects will present artifacts under some environment lights owing to imperfect geometry.
As shown in Figure~\ref{fig:limitation}, there exist some weird textures on the top or the side of the bread, and the relit strawberry presents many undesirable hollows.
The problem of imperfect geometry is not our focus here and is expected to be solved in future research.
Besides, 3D assets generated by our current method still lack diversity and we are interested in circumventing it with promising Variational Score Distillation (VSD) loss proposed in~\cite{wang2023prolificdreamer}.